\documentclass{bmvc2k}

\title{Directed-Weighting Group Lasso\\ For Eltwise Blocked CNN Pruning}

\addauthor{Ke Zhan}{zhanke@emails.bjut.edu.cn}{1}
\addauthor{Shimiao Jiang}{shimiao.jsm@alibaba-inc.com}{2}
\addauthor{Yu Bai}{bysowhat@gmail.com}{3}
\addauthor{Yi Li}{liyi30@jd.com}{3}
\addauthor{Xu Liu}{liuxu6@jd.com}{3}
\addauthor{Zhuoran Xu}{xuzhuoran@jd.com}{3}

\addinstitution{
 Faculty of Information Technology\\
 Beijing University of Technology\\
 Beijing, China
}
\addinstitution{
 Alibaba, Inc.\\
 No. 10, Furong Street, Chaoyang District,\\
 Beijing, China
}
\addinstitution{
 JD.com, Inc.\\
 No. 18, Kechuang 11th Street, Yizhuang Economic and Technological Development Zone, \\
 Beijing, China
}

\runninghead{Student, Prof, Collaborator}{BMVC Author Guidelines}

\def\etal{\emph{et al}\bmvaOneDot}

\begin{document}

\maketitle

\begin{abstract}
Eltwise layer is a commonly used structure in the multi-branch deep learning network. In a filter-wise pruning procedure, due to the specific operation of the eltwise layer, all its previous convolutional layers should vote for which filters by index should be pruned. Since only an intersection of the voted filters is pruned, the compression rate is limited. This work proposes a method called Directed-Weighting Group Lasso (DWGL), which enforces an index-wise incremental (directed) coefficient on the filter-level group lasso items, so that the low index filters getting high activation tend to be kept while the high index ones tend to be pruned. When using DWGL, much fewer filters are retained during the voting process and the compression rate can be boosted. The paper test the proposed method on the ResNet series networks. On CIFAR-10, it achieved a 75.34\% compression rate on ResNet-56 with a 0.94\% error increment, and a 52.06\% compression rate on ResNet-20 with a 0.72\% error increment. On ImageNet, it achieved a 53\% compression rate with ResNet-50 with a 0.6\% error increment, speeding up the network by 2.23 times. Furthermore, it achieved a 75\% compression rate on ResNet-50 with a 1.2\% error increment, speeding up the network by 4 times.
\end{abstract}

\section{Introduction}
\label{sec:intro}

Deep convolutional neural networks (CNNs) have recently achieved great success in many visual recognition tasks.  Many works have been proposed to compress large CNNs or directly learn more efficient CNN models for fast inference. However, the eltwise layer's eltwise operation largely limits the network pruning ability. It is one of the important problems that need to be solved and also the purpose of this research.

\section{Related Work}
\label{sec:relatedwork}

Among deep neural network model compression and acceleration, parametric pruning is one of the commonly used methods. It achieves model compression by removing redundant and less informed weights from the trained network \cite{RN1}. It is robust to various settings, can achieve good performance, and support both training from scratch and pre-trained models \cite{RN1}. Srinivas and Babu had proposed a neuron-oriented pruning method by studying the redundancy between neurons \cite{RN2}. Han \etal had proposed the classic train-prune-finetuning three-step pruning method, which used L2-norm for neuron-level pruning \cite{RN3}. Wen \etal had proposed a Structural Sparse Learning (SSL) model which enforced group lasso constraints on multiple levels CNN structures like filter and channel before performing sparse learning and pruning \cite{RN4}. Li \etal used L1-norm for filter-level pruning, they pointed out that the L1-norm and L2-norm mechanisms are different while the effects are similar \cite{RN5}.

CNN network with branch structure usually brings better performance, and convolution-eltwise is a commonly used module in branched CNN. Due to the specific operation of the eltwise layer, all its previous convolutional layers should vote for which filters by index should be pruned. Since only an intersection of the votes is pruned, the pruning ability is limited. One typical branched CNN is the ResNet \cite{RN6}. Many pruning studies use the ResNet as an example to verify the effectiveness of their pruning methodology. One consensus has been reached that it is difficult to deeply prune ResNet \cite{RN7, RN8}. There are many works that have studied ResNet pruning. Wu \etal proposed the SqueenzeDet \cite{RN9}. Huang \etal proposed a random depth network pruning method \cite{RN10}. Ullrich \etal proposed a simple regularization method based on soft weight-sharing, which included both quantization and pruning in one simple (re-)training procedure \cite{RN8}. Wang \etal used successive incremental coefficients for group lasso items in different fine tuning iterations \cite{RN11}. The method addressed the problem that ResNet's initial pruning cannot withstand large penalty items. Ding \etal proposed the AFP method which divided different filters of the same convolutional layer into two groups named reserved/removed \cite{RN12}. AFP assigned different coefficients on L1-norm of filters during training and later pruned the removed group. Our method assigns different coefficients for different group lasso items, and it is similar to \cite{RN11} and \cite{RN12}. The differences lay in the problem to be solved and the design of coefficient setting. Our method assigns coefficients using an index related function for each filter of the same convolutional layer in one fine-tuning round, and its purpose is to increase the compression rate of CNNs having convolution-eltwise modules like ResNet.

\section{Problem Definition}
\label{sec:problemdefinition}

Following the definition of group lasso constraints on CNN in \cite{RN4}, set $W_k^l$ as the k-th filter, $E(W)$ as the loss on data, $R(W)$ as non-structured regularization (e.g. l2-norm), $R_G (W_k^l)$ as filter-wise group lasso. $\lambda$ and $\lambda_G$ are empirical coefficients on corresponding items. The generic optimization target is shown as formula 1. $R_G (W_k^l)$ is the square root of the sum of j-th weight $w_(k,j)^l$ in filter k of layer l, as shown in formula 2. Group lasso can effectively zero out all weights in groups, thus filter-wise group lasso would zero out a whole filter and then the filter is removed from the convolution layer. This work also uses formula 2 as the measure of filter activation.

$$E(W)=E(W)+\lambda R(W)+\lambda_G\sum_{l=1}^L\sum_{k=1}^K R_G (W_k^l) \eqno{(1)}$$
$$R_G (W_k^l)=\sqrt{(\sum_{j=1}^{\|w_k^l\|} {(w_{k,j}^l)^2 }} \eqno{(2)}$$

Here we further assume that one CNN contains at least one eltwise layer (ELT) and an ELT has $N$ convolution layers ($Conv=\{Conv_1,\ldots,Conv_N\}$) as its direct/indirect bottom layers. In a usual pruning scenario, $N$ convolutional layers are pruned separately. Each of these convolutional layers has $F$ filters, indexed from $1$ to $F$. For $Conv_i$, the set of the indices of the pruned filters is named as $RM_i$.

For pruning filters of a convolutional layer before ELT, all $Conv_i$ should vote for which filters by index to be pruned. Two voting strategies exist: 1) prune the intersection as $RM=\cap_i\{RM_i\}$  or 2) prune the union as $RM=\cup_i\{RM_i\}$ . The former can better preserve the precision, while it is difficult to reach a high compression rate; the latter can improve the clipping amount better, while the precision drop is large and hard to restore as many important activated filters are removed.

\begin{figure}
\begin{center}
\begin{tabular}{c c}
\bmvaHangBox{\fbox{\includegraphics[width=5.3cm]{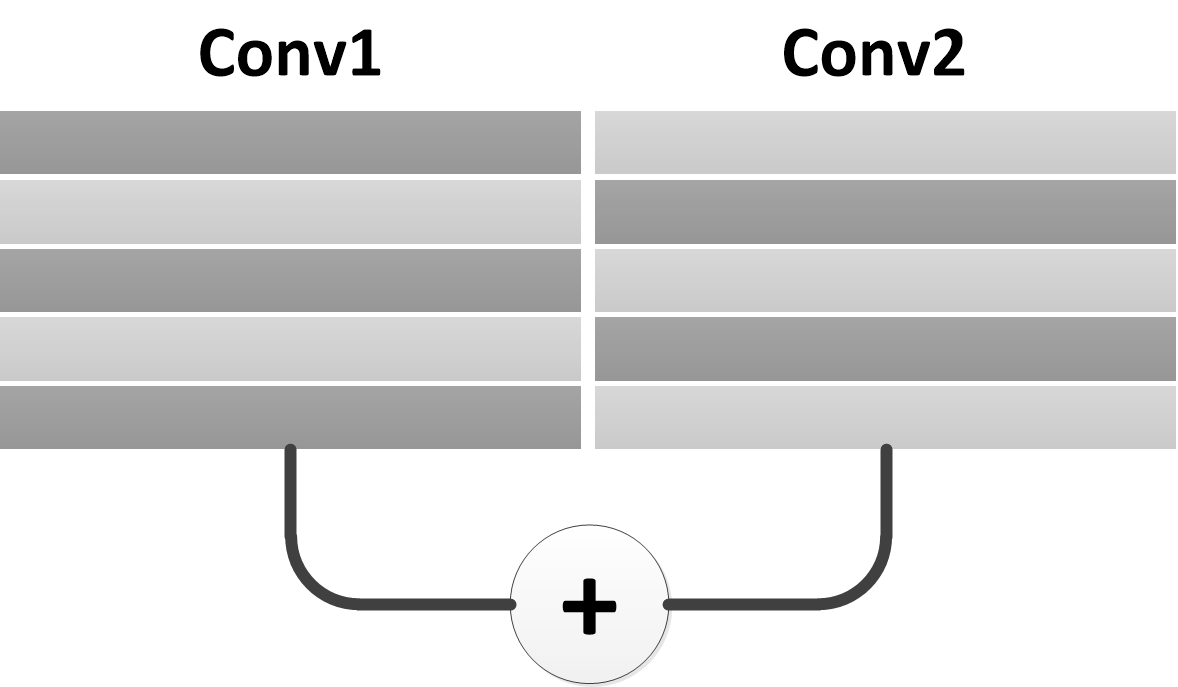}}}&
\bmvaHangBox{\fbox{\includegraphics[width=5.3cm]{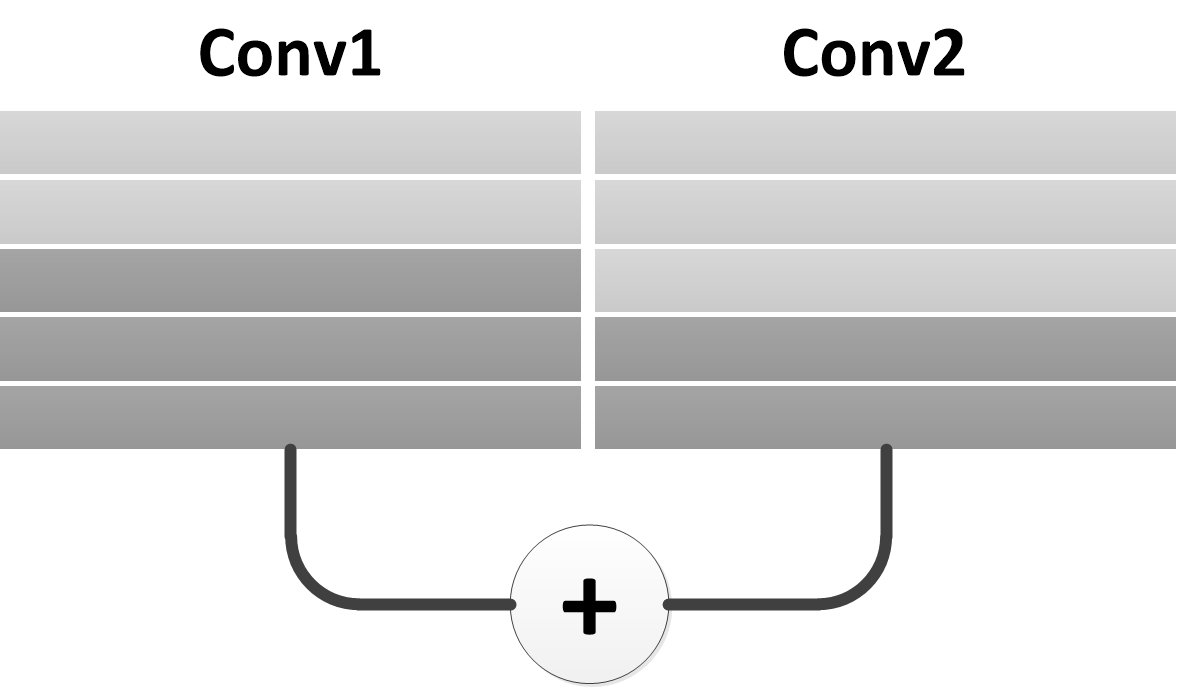}}}\\
(a)&(b)
\end{tabular}
\end{center}
\caption{ Activate Mode: (a)Filters Activated Randomly; (b) Filters Activated Under Control.}
\label{fig:teaser}
\end{figure}

As shown in Figure 1 (A), as Conv1 will pruning filter $\{f2, f4\}$, Conv2 prefers pruning $\{f1, f3, f5\}$. Strategy 1 will pruning nothing while strategy 2 will loss all its precision. The disadvantage is brought by filter randomly activation. If filters of each Conv are incremental activated, the problem would be solved. As shown in Figure 1(B), if Conv1 prefers prune filter $\{f1, f2, f3\}$ and Conv2 prefers pruning of $\{f1, f2\}$. Combining strategy 1, $\{f1, f2\}$ are pruned, the compression rate is $40\%$ as while as the max precision is kept. Combining strategy 2, $\{f1, f2, f3\}$ are pruned, the compression rate is as high as $60\%$.

\section{Methodology}
\label{sec:methodology}

To solve the problem brought by randomly activated filters, this work proposes a method to replace random filter activation by controlled filter activation through index-wise filter weighting, called Directed-Weighting Group Lasso (DWGL). It assigns a small coefficient on 1-index filter's item $R_G (W_k^l)$, and gradually increase the coefficient until the last. Notion that although trends in either direction is allowed, the work uses an increment coefficient as default. The coefficient setting was outputted by a coefficient function $f(k)$, which obeys 5 basic rules as detailed in methodology section. The modified formula is shown as formula 3. For simplify, this work named the Layer L' sum of filter-wise group lasso as $T_G^l (W)$ in formula 4.

$$E(W)=E(W)+\lambda R(W)+\lambda_G\sum_{l=1}^L\sum_{k=1}^K f(k) R_G (W_k^l) \eqno{(3)}$$
$$T_G^l (W)= \sum_{k=1}^K f(k) R_G (W_k^l) \eqno{(4)}$$

The design of $f(k)$ requires the following five principles. 
\begin{enumerate}
\item Directed Trend. Function $f(k)$ should be incremental with filter index $k$, which should lead to an decremented activation.
\item Exponential growth. To guarantee the decremented trend of filter activation, value of $f(k)$ should increase exponentially with index $k$.
\item Magnitude invariance. Item $T_G^l (W)$ should keep its magnitude although the coefficient $f(k)$ is forced.
\item Fix magnitude ratio. Experience shows that the most efficient value domain of $f(k)$ is $[10^{-1}, 10^{-5}]$, when $f(k)>10^{-1}$, the filter tends to be heavily suppressed; when $f(k)<10^{-5}$, the filter tends to be loosely released. The coefficient should better adapts to this range.
\item Scale invariance. The range of $f(k)$ is not affected by the range of $k\in[1,\ldots­,K]$.
\end{enumerate}

Based on above 5 rules, the coefficient function $f(k)$ is designed as formula 5. In formula 5, the softmax style formula meets principle 1, 2 and 3. The value 9.22 scales the range of $f(k)$, which better fits principle 4. The ratio $\frac{9.22k}{K}$ guarantees the principle 5.
  
$$f(k)=\frac{e^\frac{9.22k}{K} }{\sum_{k=1}^K e^\frac{9.22k}{K}} \eqno{(5)}$$

The pruning process is consistent with the way provided by \cite{RN12}, and here will not be described again.

\section{Experiments and Result Discussion}
\label{sec:expandres}

In this section, the work first analyzes the actual impact of DWGL on one single convolutional layer, which proves that DWGL can force controlled decremental activation. Secondly, it verifies its effect by applying the Resnet-56 and Resnet-20 model on CIFAR-10 dataset and the Resnet-50 model on ImageNet dataset respectively, which proves that DWGL can release speed acceleration and good final performance.

To facilitate comparison, this work uses the same experiment configuration as those mention in \cite{RN11, RN12}. All of the following experiments use Caffe as the experimental framework. Each experiment use four Nvidia P40 GPU as the training and testing hardware.

\subsection{Effect of DWGL on Convolution}

This article compares the differences between the original ResNet, ResNet using group lasso, and ResNet using DWGL. Among them, the original ResNet only uses L2-norm to constrain the global weights; ResNet using group lasso apply filter level group lasso. Experiments in this section were carried out on the ResNet-50 network. In figure 2 and 3, the horizontal axis represents the filter index of the convolutional layer, and the vertical axis represents the filters' activation value (formula 2). As experiment result shown in Figure 2, filter activations of the original ResNet convolutional layer is high in median 1.88, low in max activated 3.77, low in variance 0.71 and rather randomized. The group lasso has the lowest median activation (1.07E-4), high in max activated 6.68, high in variance 3.06. For DWGL, the division between the activated/non-activated filters is obvious, showing a downward trend. Split by the overall average value, the number of activated filter is 132, median 3.00, variance 2.31, while the overall average activation 1.54, median 0.43. 

This paper also compares the activation pattern of conv layers with 64/512/2048 filters. As experiment result shown in Figure 3, regardless of the number of filters, the activation exhibits the same activation pattern. The low indexed filters are activated while the high indexed filters are suppressed.

The above experimental results are consistent with the design, which show DWGL produces two expected effects, 1. force the low indexed filters to be activated and the high indexed filters to be suppressed; 2. The number of filters would not affect the activation pattern.

\begin{figure}
\begin{center}
\bmvaHangBox{\fbox{\includegraphics[width=12.64cm]{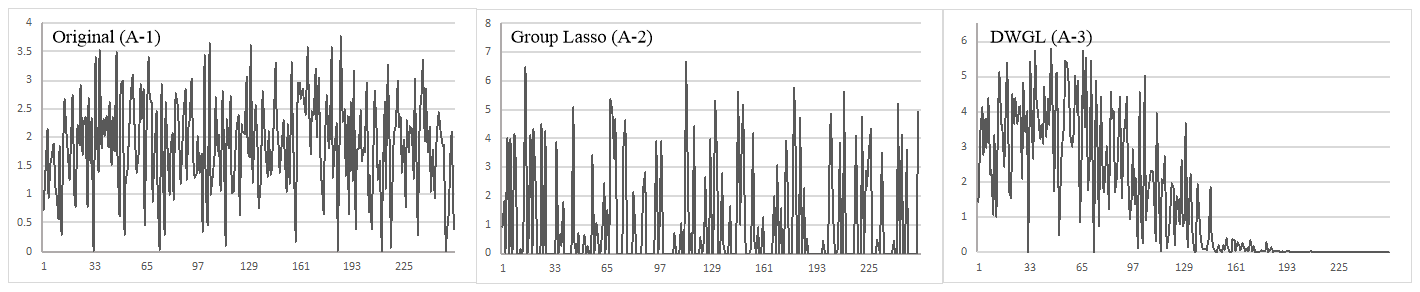}}}
\caption{Comparison of Activation Between Original, Group Lasso and DWGL.}
\end{center}
\label{fig:layeractivationcompare}
\end{figure}

\begin{figure}
\begin{center}
\bmvaHangBox{\fbox{\includegraphics[width=12.64cm]{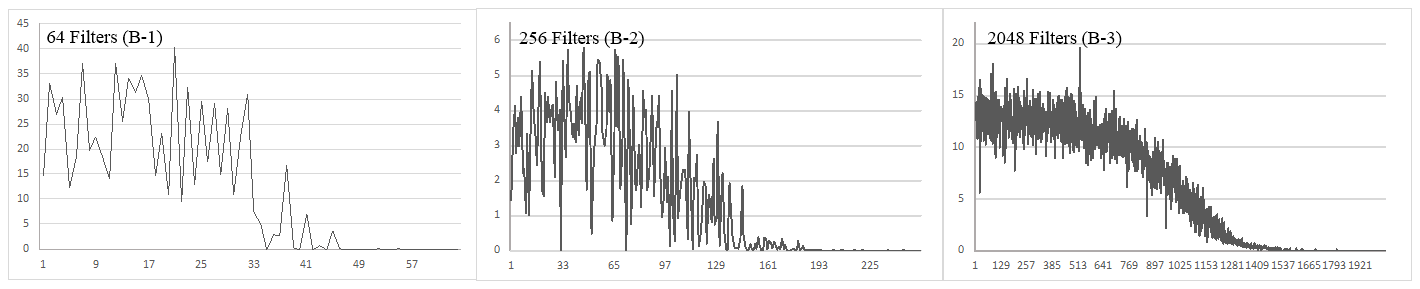}}}
\caption{Comparison of DWGL Activation of 64/256/2048 Filters.}
\end{center}
\label{fig:filteractivationcompare}
\end{figure}

\subsection{Resnet-56 / Resnet-20 on CIFAR-10 dataset}

The work retrain the ResNet-56 model accuracy to 92.99\% and retrain the ResNet-20 model accuracy to 90.9\% on CIFAR-10 dataset as the baseline, and execute DWGL on them. 
The pruning result show that, DWGL reach a high compression rate of 74.64\% with an increase error of only 0.94\%. Compared with the AFP \cite{RN12} method, DWGL increase compression rate by 14.48\%, while causing less increased error by 0.06\%. At the same time, DWGL shows more potential, when the pruning reached an increased error of 1.05\%, the compression rate could be as much as 80\%.

Comparison of DWGL and SSL \cite{RN4} is not intuitive enough, since which convolution layer was pruned and the compression rate were not reported in their work. As SSL prune 20 layers to 14 within an increased error of about 1.0\%, DWGL reached a compression rate 52.06\% with an increased error of about 0.72\%.

\begin{table}
\begin{center}
\begin{tabular}{|l|c|c|c|}
\hline
Method & Models & Compression Rate & Increase Error\% \\
\hline\hline
SSL & Resnet-20 & 20 to 16 layers & ~0.6 \\
SSL & Resnet-20 & 20 to 14 layers & ~1.0 \\
ours & Resnet-20 & 52.06\% & 0.72 \\
AFP & Resnet-56 & 60.86\% & 0.99 \\
ours & Resnet-56 & 75.34\% & 0.94 \\
ours & Resnet-56 & 80\% & 1.05 \\
\hline
\end{tabular}
\end{center}
\caption{Test Result on CIFAR-10.}
\end{table}

\subsection{ResNet-50 on ImageNet}

The work re-train the ResNet-50 model accuracy to 91.3\% as the baseline and execute DWGL on it. The pruning result show that, DWGL reach a compression rate of 53\% with an increase error of only 0.6\%, and the speed up is 2.23x. When reaching the 2x speed up, the increased error of method CP \cite{RN13}, SPP \cite{RN14} and SPIR \cite{RN11} is correspondingly 1.4\%, 0.8\% and 0.1\%. A further experiment show that the proposed DWGL model can reach a compression rate as much as 75\% and the corresponding speed up is 4x at a cost of increased error of only 1.2\%.

\begin{table}
\begin{center}
\begin{tabular}{|l|c|c|c|c|}
\hline
Method & Models & Compression Rate & Speedup Rates & Increase Error\% \\
\hline\hline
CP & Resnet-50 & - & 2x & 1.4 \\
SPP & Resnet-50 & - & 2x & 0.8 \\
SPIR & Resnet-50 & - & 2x & 0.1 \\
ours & Resnet-50 & 53\% & 2.23x & 0.6 \\
ours & Resnet-50 & 75\% & 4x & 1.2 \\
\hline
\end{tabular}
\end{center}
\caption{Test Result on ImageNet.}
\end{table}

\section{Conclusions and Future Work}
\label{sec:conclusion}

This paper proposes directed-weighting group lasso method for pruning of eltwise blocked CNNs, called DWGL. DWGL solves the pruning problem that brought by the random filter activation and eltwise operation in convolution-eltwise modules through controlled filter activation. Although reported only on ResNet, this method is widely applicable to deep CNN networks with eltwise layers. This method can raise the compression rate with little/no increased error. The method never refuses to combine other accelerating method like quantization, so the pruning performance may be further boosted. The weakness of DWGL is that, like all other parameter pruning methods, DWGL requires iteratively pruning and fine-tuning, which usually take a longer time. In the future, this issue needs continuous improvement. Since DWGL is effective and robust, how to apply its basic idea to non-CNN networks could also be further researched.

\bibliography{dwgl}

\begin{thebibliography}{14}
\providecommand{\natexlab}[1]{#1}
\providecommand{\url}[1]{\texttt{#1}}
\expandafter\ifx\csname urlstyle\endcsname\relax
  \providecommand{\doi}[1]{doi: #1}\else
  \providecommand{\doi}{doi: \begingroup \urlstyle{rm}\Url}\fi

\bibitem[Ding et~al.(2018)Ding, Ding, Han, and Tang]{RN12}
Xiaohan Ding, Guiguang Ding, Jungong Han, and Sheng Tang.
\newblock Auto-balanced filter pruning for efficient convolutional neural
  networks.
\newblock In \emph{Thirty-Second AAAI Conference on Artificial Intelligence},
  pages 6797--6804, 2018.

\bibitem[Han et~al.(2015)Han, Pool, Tran, and Dally]{RN3}
Song Han, Jeff Pool, John Tran, and William Dally.
\newblock Learning both weights and connections for efficient neural network.
\newblock In \emph{Advances in Neural Information Processing Systems}, pages
  1135--1143, 2015.

\bibitem[Hao et~al.(2016)Hao, Kadav, Durdanovic, Samet, and Graf]{RN5}
Li~Hao, Asim Kadav, Igor Durdanovic, Hanan Samet, and Hans~Peter Graf.
\newblock Pruning filters for efficient convnets.
\newblock \emph{arXiv preprint arXiv:1608.08710}, 2016.

\bibitem[He et~al.(2016)He, Zhang, Ren, and Sun]{RN6}
Kaiming He, Xiangyu Zhang, Shaoqing Ren, and Jian Sun.
\newblock Deep residual learning for image recognition.
\newblock In \emph{Proceedings of the IEEE Conference on Computer Vision and
  Pattern Recognition}, pages 770--778, 2016.

\bibitem[He et~al.(2017)He, Zhang, and Sun]{RN13}
Yihui He, Xiangyu Zhang, and Jian Sun.
\newblock Channel pruning for accelerating very deep neural networks.
\newblock In \emph{Proceedings of the IEEE International Conference on Computer
  Vision}, pages 1389--1397, 2017.

\bibitem[Huang et~al.(2016)Huang, Sun, Liu, Sedra, and Weinberger]{RN10}
Gao Huang, Yu~Sun, Zhuang Liu, Daniel Sedra, and Kilian~Q Weinberger.
\newblock Deep networks with stochastic depth.
\newblock In \emph{European conference on computer vision}, pages 646--661.
  Springer, 2016.

\bibitem[Luo and Wu(2017)]{RN7}
Jian~Hao Luo and Jianxin Wu.
\newblock An entropy-based pruning method for cnn compression.
\newblock \emph{arXiv preprint arXiv:1706.05791}, 2017.

\bibitem[Srinivas and Babu(2015)]{RN2}
Suraj Srinivas and R~Venkatesh Babu.
\newblock Data-free parameter pruning for deep neural networks.
\newblock \emph{arXiv preprint arXiv:1507.06149}, 2015.

\bibitem[Ullrich et~al.(2017)Ullrich, Meeds, and Welling]{RN8}
Karen Ullrich, Edward Meeds, and Max Welling.
\newblock Soft weight-sharing for neural network compression.
\newblock \emph{arXiv preprint arXiv:1702.04008}, 2017.

\bibitem[Wang et~al.(2017)Wang, Zhang, Wang, and Hu]{RN14}
Huan Wang, Qiming Zhang, Yuehai Wang, and Haoji Hu.
\newblock Structured probabilistic pruning for convolutional neural network
  acceleration.
\newblock \emph{arXiv preprint arXiv:1709.06994}, 2017.

\bibitem[Wang et~al.(2018)Wang, Zhang, Wang, and Hu]{RN11}
Huan Wang, Qiming Zhang, Yuehai Wang, and Haoji Hu.
\newblock Structured pruning for efficient convnets via incremental
  regularization.
\newblock \emph{arXiv preprint arXiv:1811.08390}, 2018.

\bibitem[Wen et~al.(2016)Wen, Wu, Wang, Chen, and Li]{RN4}
Wei Wen, Chunpeng Wu, Yandan Wang, Yiran Chen, and Hai Li.
\newblock Learning structured sparsity in deep neural networks.
\newblock In \emph{Advances in Neural Information Processing Systems}, pages
  2074--2082, 2016.

\bibitem[Wu et~al.(2017)Wu, Iandola, Jin, and Keutzer]{RN9}
Bichen Wu, Forrest Iandola, Peter~H. Jin, and Kurt Keutzer.
\newblock Squeezedet: Unified, small, low power fully convolutional neural
  networks for real-time object detection for autonomous driving.
\newblock In \emph{IEEE Conference on Computer Vision and Pattern Recognition
  Workshops}, pages 129--137, 2017.

\bibitem[Yu et~al.(2017)Yu, Wang, Pan, and Tao]{RN1}
Cheng Yu, Duo Wang, Zhou Pan, and Zhang Tao.
\newblock A survey of model compression and acceleration for deep neural
  networks.
\newblock \emph{arXiv preprint arXiv:1710.09282}, 2017.

\end{thebibliography}
\end{document}